# Relations World: A Possibilistic Graphical Model[1]

Christopher J.C. Burges, Erin Renshaw, and Andrzej Pastusiak

Microsoft Research

November 14th, 2014

## Abstract

We explore the idea of using a *possibilistic graphical model* as the basis for a world model that drives a dialog system. As a first step we have developed a system that uses text-based dialog to derive a model of the user's family relations. The system leverages its world model to infer relational triples, to learn to recover from upstream coreference resolution errors and ambiguities, and to learn context-dependent paraphrase models. We also explore some theoretical aspects of the underlying graphical model.

## Introduction

Recently we listed some desiderata that appear to us to be sensible design principles in the quest to develop the machine comprehension of text [1]. We summarize them here:

1. Inference should leverage <u>world models</u> which are kept
2. as <u>simple</u> as possible such that
3. they can be combined in a <u>modular</u> fashion. This should help us keep the design
4. <u>scalable</u>, in three senses: learning time, inference time, and portability to arbitrary domains. Learning should be
5. <u>correctable</u>, so that mistakes the system makes can be corrected without introducing new mistakes elsewhere,
6. <u>interpretable</u>, so that for example developers can easily understand what inferences the system has made, and all the reasoning it has done,
7. <u>generative</u>, so that the system can ask the user for more information about the data it would benefit most from knowing, and
8. <u>interrogable</u>, so that the user can ask if the system believes something and get an answer in real time.

(1) and (2) are uncontroversial: a world model is often needed to resolve ambiguities in language. For example, in the sentence *Sam is my father and I have a brother named Sam*, a family relations model is needed to infer that the two Sams must refer to different entities. Regarding (3), modularity, and composability of those modules, is a fundamental principle of software design but is also expected to be a key design principle for large semantic models [2]. (4), especially portability, is an open research question. (5) is a key property that distinguishes human learning from machine learning (ML). Most ML approaches do not address this issue at all. Errors are corrected by adding more training data, or by designing better features, or by finding a better model, but all of these approaches will in general introduce new errors on the original data; a human analog would be that one consequence of learning to ride a bicycle is that a child forgets how to brush his teeth. In this sense (3) is not addressed by ML either: a human analog would be a teacher correcting a student's misunderstanding by locking him in a library with terabytes of data for a week and asking him to update his parameters. Instead, the teacher can correct the student's misunderstanding by exchanging a small number of

---

bits of information, correcting a specific component (module) of the student's world model. It seems unlikely that the brain is moving global decision surfaces around in some high dimensional space. This also argues strongly for (1), i.e. a rich shared world model, which even allows entities to model each others' models. ML models (neural nets, trees, ensembles of these, and even moderately sized probabilistic graphical models) are not easily interpretable (see for example [3]): in general we do not understand why our statistical systems make the errors they do. ML achieves (7) to some extent through active learning, but not in the sense of maximally reducing the uncertainty in a semantic world model. Similarly while ML approaches (8) by simply applying the model to unseen data, it does not do so in a modular fashion, so that for example there are no semantic components in the model that we can interrogate individually.

The approach described here, although only a preliminary step, meets all these desiderata except for (4), which we hope to investigate as the models grow in complexity. We call our simple world models *meaning projections,* where a given meaning projection extracts meaning about one, or a very few, aspects of a piece of text (for example, spatial information about the entities involved), and can interact with other meaning projections to both correct upstream errors (for example, coreference resolution) and to make its own inferences. The idea is roughly analogous to the "agents" in Minsky's Society of Mind [4]. Our first step, as outlined in this paper, is a dialog system[2] that maps English text to model family relations. The work can be viewed as a semantic modeling analog of Blocks World [4,5], in that it takes a simple task and uses it to explore the ideas. Although the machine is learning, there is currently no statistical machine learning involved; rather than using probabilistic graphical models, uncertainties are modeled using graphs and sets, and robustness is achieved by carefully tracking the sources of the uncertainty and polling the user directly when necessary. Clearly statistical methods will need to be added at some point, and one goal of this research is to understand how to incorporate them while meeting the above desiderata. Here, we concentrate on an approach called *possibilistic graphical models*[3] (pGMs) that are a simple instantiation of classical relational logic [6], but with data structures chosen to meet the above desiderata. The uncertainty in the model is exposed explicitly and actionably. While logic problems involving family relations have been given as homework exercises since the beginning of AI, we've found it a useful test bed to focus on, in particular to generate dialog driven by the uncertainty in the model, and to explore growing and correcting the model through dialog.

## A Possibilistic Graphical Model

We explain pGMs by describing how we use them to model family relations. Assume that we are given a set of $n$ animate entities ("AEs") (usually, but not necessarily, people[4]), denoted below by $E_i, i = 1, ..., n$. Each AE has an associated name and gender which may or may not be known. The AEs are not necessarily unique: for example, upstream natural language processing may have in error separated entities which are in fact the same. It is also possible that what is presented to the system as a single AE may in fact refer to more than one, due to upstream coreference resolution errors: the system can also (often) detect and correct these kinds of error. The pGM is a directed multigraph with a node for each AE and an edge set for the possible relations between those two AEs. Thus an edge set represents our current state of knowledge about the possible relations that could hold for the two endpoints: it is assumed that exactly one of the relations is correct, but we do not know which. The pGM grows as more entities are encountered in the text, and its edge sets shrink (to a minimum cardinality of one) as more information concerning its entities is discovered. Our pGM is populated from binary predicates (we use the term "relations" interchangeably), for example *Mother(Anne,Bill)*, derived from natural text using a few simple pattern matching rules; from unary predicates (e.g. *IsFemale(Anne)*) derived using Census data and user input; and consistency rules, such as gender propagation (where for example *spouseOf* would be resolved to *wifeOf* if its first argument is discovered to be female[5]; the second argument would be set to *male;* the outgoing relations of the latter might then be resolved from *parentOf* to *fatherOf*; and so on). While this graph constitutes our evolving knowledge base, we also use a (hand crafted) fixed knowledge base matrix $M$ that captures relations between relations. Both rows and columns of $M$ are indexed by a relation, which is a member of the fixed set of relations $\mathbb{R}$ that we wish to model. Our current system has $\mathbb{R} = \{$*Grandparent, Parent, Parent-in-Law, Spouse, Sibling, Sibling-in-Law, Child, Child-in-Law, Grandchild, Aunt or Uncle, Niece or Nephew, Cousin, Self, OutOfGraph*$\}$ together with the gender specific versions of these. We will denote any subset of $\mathbb{R}$ by $\mathcal{R}$ and an individual element by $R$. *Self* is used to model the fact that some entities that are passed to the system as different may in reality be the same (thus enabling the system to correct coreference errors). *OutOfGraph,* which we denote below by $R_0$, is necessary to model the fact that we know that an induced relation may not be in the fixed knowledge base.

---

[2] A demo of the system will be presented at the workshop.

[3] We use pGM to emphasize that possibilistic graphical models are precursors to probabilistic GMs (PGMs).

[4] For example when modeling fiction, one might have to model the fact that objects which in the real world are inanimate are treated in the text as though they are animate, with gender, name, mood, etc.

The graph encodes relations as follows. Let $\mathcal{R}_{ij}$ be a set of relations such that for exactly one element $R \in \mathcal{R}_{ij}$, we have that $R(E_i, E_j) = T$; we do not know which element this is, unless $|\mathcal{R}_{ij}| = 1$. These sets of relations are captured in a matrix $M$ such that $M_{R_1 R_2} = \mathcal{R}^{12}$, which we also write, for any entities $E_1, E_2, E_3$, as

$$R^{(1)}(E_1, E_2) \wedge R^{(2)}(E_2, E_3) \rightarrow \mathcal{R}^{(12)}(E_1, E_3) \tag{1}$$

Sometimes the set $\mathcal{R}^{(12)}$ will be a singleton (e.g. $\texttt{Parent}(A,B) \wedge \texttt{Sibling}(B,C) \rightarrow \texttt{Parent}(A,C)$), but it need not be (e.g. $\texttt{Cousin}(A,B) \wedge \texttt{Cousin}(B,C) \rightarrow \mathcal{R}(A,C)$ where $\mathcal{R} = \{\texttt{Cousin}, \texttt{Self}, \texttt{Sibling}, R_0\}$). Every relation is assumed to have a unique inverse: that is, for any $E_1$ and $E_2$, and for every predicate $R$, there is a unique predicate $R^{-1} : R(E_1, E_2) \Leftrightarrow R^{-1}(E_2, E_1)$. Thus (1) also implies the set $\mathcal{R}^{(21)}(E_3, E_1)$ which is in 1-1 correspondence with the set $\mathcal{R}^{(12)}(E_1, E_3)$ and where each element of $\mathcal{R}^{(21)}$ is the inverse of an element in $\mathcal{R}^{(12)}$. We use $\mathcal{R}_{ij}$ to denote the set of edges from node $E_i$ to node $E_j$ and $\mathcal{R}_{ji}$ to denote the corresponding set of inverse edges. Equation (1) may also be written with sets on the left, in which case the right hand side is the union of all such implications: that is, we use $M(\mathcal{R}_{ij}, \mathcal{R}_{jk})$ as shorthand for $\bigcup_{R_1 \in \mathcal{R}_{ij}, R_2 \in \mathcal{R}_{jk}} M(R_1, R_2)$.

Graphs are always either fully connected, or are composed of fully connected subgraphs. There are two kinds of updates. The first takes an existing graph, updates one of the edges by reducing its multiplicity (to at least one), and propagates the inferred edge set reductions, recursively. Thus suppose that we have a 3-clique[6] $\mathcal{R}_{ij}, \mathcal{R}_{jk}, \mathcal{R}_{ik}$, and that new information leads us to replace $\mathcal{R}_{ij}$ by $\mathcal{R}'_{ij} \subset \mathcal{R}_{ij}$. For example, if $\mathcal{R}_{12} = \{\texttt{Cousin}, \texttt{Sibling}, \texttt{Self}, R_0\}$, then one initial update might be that we have learned that $E_1$ is not the self of $E_2$. The first step of the recursion is then either to replace $\mathcal{R}_{ik}$ by $\mathcal{R}'_{ik} := M(\mathcal{R}'_{ij}, \mathcal{R}_{jk}) \cap \mathcal{R}_{ik}$, or to replace $\mathcal{R}_{jk}$ by $\mathcal{R}'_{jk} := M(\mathcal{R}_{ji}, \mathcal{R}_{ik}) \cap \mathcal{R}_{jk}$ (we prove below that it doesn't matter which we do first). The second kind of update is when an edge is added to join previously disconnected graphs, which after applying inference results in a single fully connected graph. One can think of any relation as always being added or removed simultaneously with its inverse relation.

---

[5] The system currently models "traditional" family structures to keep the analysis simple. Relations resulting from, for example, same sex marriages, or step-relations, are easily added.

[6] We use the term "3-clique" rather than "triangle" because the structure is more general than a geometric construct.

We emphasize that inference in this system is simply using relational logic [6]. Our intent here is to examine data structures that support the above desiderata, in particular correctability and interpretability, through the ability to track uncertainty explicitly, efficiently and actionably. Thus, for example, a pGM yields all possible relations between any two entities by simply examining the edge set connecting those two entities. To this end we provide some analysis of pGMs here.

**Definition 1:** We overload the $\rightarrow$ symbol as follows: for any two edges $R_1, R_2$ from different edge sets and which join at node $E$ in a 3-clique, we write $R_1 \rightarrow R_2$ to mean that there exists an $R$ from the third edge set (or its inverse) such that the following holds: for $R_1, R_2$ both terminating on $E$, $R_2 \in M(R, R_1)$; for both $R_1$ and $R_2$ starting at $E$, $R_2 \in M(R_1, R)$; for $R_1$ ending at $E$ and $R_2$ starting there, $R_2 \in M(R_1^{-1}, R)$; and for $R_1$ starting at $E$ and $R_2$ ending there, $R_2 \in M(R, R_1^{-1})$. We read $R_1 \rightarrow R_2$ as "$R_1$ supports $R_2$" and we define $R \rightarrow \mathcal{R}$ to mean that every edge in $\mathcal{R}$ is supported by $R$, and $\mathcal{R}_1 \rightarrow \mathcal{R}_2$ to mean that every edge in $\mathcal{R}_2$ is supported by at least one edge in $\mathcal{R}_1$.

Note that the empty set is supported by any pair of relations. Note also that if $R$ is supported by another edge in a 3-clique, it does not mean that $R$ must occur, just that its occurrence is not inconsistent with the other edges in the clique.

**Definition 2:** A 3-clique with (different) vertex indices $a, b, c \in \mathbb{N}$ is *stable* if $R_{ij} \subseteq M(R_{ik}, R_{kj}) \ \forall \{i, j, k\} \in PRM\{a, b, c\}$, that is, if every edge set is supported by the other two. A fully-connected graph is stable if all of its three-cliques are stable.

Note that, given the definition of the inverse of an edge, and since $M$ is just the encapsulation of the logical assertion (1), we have that $M(R_1, R_2) = \mathcal{R}_3 \Leftrightarrow M(R_2^{-1}, R_1^{-1}) = \mathcal{R}_3^{-1}$. That is, for predicates $p_i$, if $p_1 \wedge p_2 \Rightarrow p_3 \vee p_4 \vee \ldots \vee p_n$ and if $p_i \Leftrightarrow q_i$ then $q_1 \wedge q_2 \Rightarrow q_3 \vee q_4 \vee \ldots \vee q_n$, where here the predicates $p$ are the relations and the predicates $q$, their inverses. Note also that since an edge set $\mathcal{R}_{ij}$ is supported by $\mathcal{R}_{ik}$ and $\mathcal{R}_{kj}$ iff $\mathcal{R}_{ji}$ is supported by $\mathcal{R}_{jk}$ and $\mathcal{R}_{ki}$, we can compute support by going either way around the clique.

**Axioms:**

A1. No element of $M$ is the empty set (every pair of edges has implications).

A2. Every individual relation, including $R_0$, has a unique inverse.

A3. Consistency conditions: let us assume that $M(R_1, R_2) = \mathcal{R}_3$. From A1, $\mathcal{R}_3$ is non-empty. Then:

    a. $\forall R \in \mathcal{R}_3, \ R_1 \in M(R, R_2^{-1})$ and $R_2 \in M(R_1^{-1}, R)$.

    b. $\forall R^{-1} \in \mathcal{R}_3^{-1}, \ R_2^{-1} \in M(R^{-1}, R_1)$ and $R_1^{-1} \in M(R_2, R^{-1})$.

A3 (a) is a consistency condition because if it were not so, then given $R_1$ and $R_2$, the resulting set $M(R_1, R_2)$ could be reduced to a single edge (for example, by the addition of new information) that implies the absence of $R_1$ or of $R_2$. Similarly, if A3 (b) were not so, then the edge set $M(R_2^{-1}, R_1^{-1})$ could be reduced to a single edge that implied the removal of $R_1^{-1}$ or $R_2^{-1}$.

**Lemma 1:** Joining nodes 1 and 3 by adding an edge set $\mathcal{R}_{13}$, computed as $\mathcal{R}_{12} \wedge \mathcal{R}_{23}$, does not have the consequence of reducing the existing edge sets $\mathcal{R}_{12}$ and $\mathcal{R}_{23}$.
**Proof:** This follows directly from axiom A3. □

**Lemma 2:** For a given 3-clique, suppose that $R_1$ belongs to one edge set and $R_2$ to another. Then $R_1 \to R_2$ if and only if $R_2 \to R_1$.
**Proof:** Clearly $R_1$ and $R_2$ belong to adjacent edge sets. Let their mutual node by $A$. There are three possible situations: $R_1$ and $R_2$ both end on $A$, $R_1$ and $R_2$ both start at $A$, or one starts and one ends on $A$ (these two are made equivalent by relabeling). If both end on $A$, then $R_1 \to R_2$ means that there exists an $R_3$ on the remaining edge such that $R_2 \in M(R_3, R_1)$. By A3 we have that $R_1 \in M(R_3^{-1}, R_2)$ so $R_2 \to R_1$. Symmetry and swapping labels $1 \leftrightarrow 2$ shows that similarly, $R_2 \to R_1$ implies $R_1 \to R_2$. If both start on A, then $(R_1 \to R_2) \Rightarrow \exists R_3 : R_2 \in M(R_1, R_3)$ and by A3, $R_1 \in M(R_2, R_3^{-1})$ or $R_2 \to R_1$. Symmetry again gives the converse result. Finally suppose that $R_1$ ends on $A$ and $R_2$ begins on $A$. Then $(R_1 \to R_2) \Rightarrow \exists R_3 : R_2 \in M(R_1^{-1}, R_3)$ and by A3, $R_1^{-1} \in M(R_2, R_3^{-1})$ or $R_2 \to R_1$ since $R_1$ has unique inverse that also holds. Similarly $(R_2 \to R_1) \Rightarrow \exists R_3 : R_1 \in M(R_3, R_2^{-1})$ and by A3, $R_2^{-1} \in M(R_3^{-1}, R_1)$ so $R_1 \to R_2$ . □

Note that Lemma 2 does not assume that the clique is stable.

The $\to$ relation is thus symmetric at the single edge level: support is mutual. A weaker (non-symmetric) condition holds for the mapping from an edge to an edge set:

**Lemma 3:** For a 3-clique with edge set $\mathcal{R}_{ij}$, let $R \in \mathcal{R}_{ij}$. Then $R \to \mathcal{R}$ implies $\mathcal{R} \to R$.
**Proof:** $R_1 \to \mathcal{R}$ means that $\forall R_2 \in \mathcal{R}, R_1 \to R_2$. By lemma 2 we have that $R_2 \to R_1 \; \forall R_2 \in \mathcal{R}$ which is also written as $\mathcal{R} \to R$. □

The assertions in Lemmas 2 and 3 do not extend to general subsets. For example, for edge $R$ and edge set $\mathcal{R}$, it is not necessarily true that $\mathcal{R} \to R$ implies that $R \to \mathcal{R}$ since $\mathcal{R}$ may be a superset of the edges supported by $R$. A fortiori, for two edge sets $\mathcal{R}$ and $\mathcal{S}$, $\mathcal{R} \to \mathcal{S}$ does not necessarily imply that $\mathcal{S} \to \mathcal{R}$.

The lemmas and theorems below assume that we start with a stable 3-clique.

**Lemma 4:** Suppose that a 3-clique has some edge set $\mathcal{R}$ with subset $\mathcal{R}'$, and another edge set $\mathcal{S}$ with subset $\mathcal{S}'$. Then removal of $\mathcal{R}'$ implies removal of $\mathcal{S}'$ if and only if $\mathcal{S}'$ only supports edges in $\mathcal{R}'$.
**Proof:** If $\mathcal{S}'$ supports only edges in $\mathcal{R}'$, then it does not support any edges in $\mathcal{R} \setminus \mathcal{R}'$ and hence, since support is mutual, $\mathcal{S}'$ is not supported by any edges in $\mathcal{R} \setminus \mathcal{R}'$; hence removal of $\mathcal{R}'$ implies removal of $\mathcal{S}'$. If on the other hand the removal of $\mathcal{R}'$ implies the removal of $\mathcal{S}'$, then there are no edges in $\mathcal{R} \setminus \mathcal{R}'$ that support $\mathcal{S}'$, and since support is mutual there are no edges in $\mathcal{S}'$ that support $\mathcal{R} \setminus \mathcal{R}'$, that is, $\mathcal{S}'$ only supports edges in $\mathcal{R}'$. □

Denoting the removal of some set $\mathcal{S}$ by $rem(\mathcal{S})$, again the generalization to arbitrary sets does not hold for removal: that is, given $rem(\mathcal{S}) \to rem(R_1)$, it may not be the case that $rem(R_1) \to rem(\mathcal{S})$ since $\mathcal{S}$ may contain elements not implied by $R_1$.

**Lemma 5:** Suppose that removal of $R \in \mathcal{R}_{ij}$ implies the removal of $\mathcal{S}$: that is, $\forall\, S \in \mathcal{S}, S \in M(\mathcal{R}_{ij}, \mathcal{R}_{jk})$ and $S \notin M(\mathcal{R}_{ij} \setminus R,\ \mathcal{R}_{jk})$. This alone cannot result in the removal of another edge $R' \in \mathcal{R}_{ij}$.

**Proof**: By Lemma 4, removal of $\mathcal{S}$ can result in the removal of $R'$ only if $R'$ supports some edges in $\mathcal{S}$. But if so then $\exists S \in \mathcal{S} : S \in M(\mathcal{R}_{ij} \setminus R, \mathcal{R}_{jk})$. □

**Lemma 6:** Suppose that, for a given 3-clique $C$ with nodes $\{i, j, k\}$, and given arbitrary edge sets $\mathcal{R}_{ij}$ and $\mathcal{R}_{jk}$, we set $\mathcal{R}_{ik} = M(\mathcal{R}_{ij}, \mathcal{R}_{jk})$. Then all resulting implications in $C$ result in no changes to any edge in $C$.

**Proof:** Consider the set $\mathcal{S}_{ij} \coloneqq M(R_{ik}, R_{kj})$. There can be no edge $R \in \mathcal{R}_{ij}$ such that $R \notin \mathcal{S}_{ij}$ since by Axiom 1, such an $R$ would support at least one relation in $\mathcal{R}_{ik}$, and by Lemma 2, that relation supports $R$; hence $\mathcal{R}_{ij} \subseteq M(R_{ik}, R_{kj})$. Similarly consider $\mathcal{S}_{jk} \coloneqq M(\mathcal{R}_{ji}, \mathcal{R}_{ik})$. There can be no edge $R \in \mathcal{R}_{jk}$ such that $R \notin \mathcal{S}_{jk}$ because by Axioms 1 and 2, $R^{-1}$ would support at least one relation in $\mathcal{R}_{ki}$, and by Lemma 2, that relation would accept $R^{-1}$; hence $\mathcal{R}_{jk} \subseteq M(\mathcal{R}_{ji}, \mathcal{R}_{ik})$. □

We now address the question of efficiently computing the full set of resultant edges in a clique when some edges are removed (due, for example, to the availability of new information). The figure below shows that reductions in one set of edges can induce reductions in another, which can then induce reductions in the third: can this process continue? (In other words, must we recurse to completion?) The following theorem shows that the answer is no.

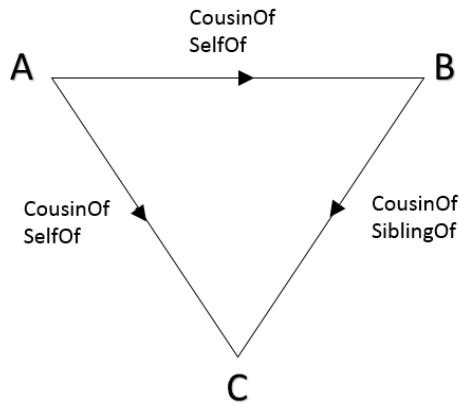

*Figure 1. Suppose that new information leads us to replace $\mathcal{R}_{AB}$ by $\mathcal{R}'_{AB} \coloneqq SelfOf$. Then the relations $\mathcal{R}'_{AB}$ and $\mathcal{R}_{BC}$ imply the reduction of $\mathcal{R}_{AC}$ to $\mathcal{R}'_{AC} \coloneqq \{CousinOf\}$. Secondly, the relations $\mathcal{R}'_{BA}$ and $\mathcal{R}'_{AC}$ imply the reduction of $\mathcal{R}_{BC}$ to $\{CousinOf\}$. In general, for arbitrary edge sets, is a continued recursion necessary? Theorem 1 shows that it is not.*

**Theorem 1:** Given a stable 3-clique, represent its edge sets by $\mathcal{R}_{ij}, \mathcal{R}_{jk}, \mathcal{R}_{ik}$. Suppose that one or more edges are removed from $\mathcal{R}_{ij}$ to give a new edge set $\mathcal{R}'_{ij}$. Suppose that $\mathcal{R}_{ik}$ is updated via $\mathcal{R}'_{ik} = M(\mathcal{R}'_{ij}, \mathcal{R}_{jk}) \cap \mathcal{R}_{ik}$, and that $\mathcal{R}_{jk}$ is then updated via $\mathcal{R}'_{jk} = M(\mathcal{R}'_{ji}, \mathcal{R}'_{ik}) \cap \mathcal{R}_{jk}$. Then these changes in $\mathcal{R}_{jk}$ and $\mathcal{R}_{ik}$ do not induce any further changes in the clique, the resulting clique is stable, and the result does not depend on the order of operations (that is, the result is the same if $\mathcal{R}_{jk}$ is updated first and $\mathcal{R}_{ik}$ second).

**Proof:** Let $\mathcal{S}_{ij} = \mathcal{R}_{ij} \setminus \mathcal{R}'_{ij}$.denote the set of edges removed from $\mathcal{R}_{ij}$. Let $\mathcal{S}_{ik}$ denote the set of edges that are not supported by $\mathcal{R}'_{ij}$ (that is, $\mathcal{S}_{ik}$ is the maximal set such that $\mathcal{S}_{ik} \subset \mathcal{R}_{ik}$ and $\mathcal{S}_{ik} \cap M(\mathcal{R}'_{ij}, \mathcal{R}_{jk}) = \emptyset$) so that the removal of $\mathcal{S}_{ij}$ implies the removal of $\mathcal{S}_{ik}$. Similarly let $\mathcal{S}_{jk}$ denote the set of edges that are simultaneously not supported by $\mathcal{R}'_{ij}$ (that is, $\mathcal{S}_{jk}$ is the maximal set such that $\mathcal{S}_{jk} \subset \mathcal{R}_{jk}$ and $\mathcal{S}_{jk} \cap M(\mathcal{R}'_{ji}, \mathcal{R}_{ik}) = \emptyset$) so that the removal of $\mathcal{S}_{ij}$ implies the removal of $\mathcal{S}_{jk}$. Also define $\mathcal{R}'_{jk} \coloneqq \mathcal{R}_{jk} \setminus \mathcal{S}_{jk}$ and $\mathcal{R}'_{ik} \coloneqq \mathcal{R}_{ik} \setminus \mathcal{S}_{ik}$. Now suppose that there were an edge $R \in \mathcal{R}'_{jk}$ whose removal is implied by the reduced set $\mathcal{R}'_{ik}$, that is, $R \notin M(\mathcal{R}'_{ji}, \mathcal{R}'_{ik})$. Then since $R \in M(\mathcal{R}'_{ji}, \mathcal{R}_{ik})$ we must have that $R \in M(\mathcal{R}'_{ji}, \mathcal{S}_{ik})$, and by lemma 2, this means that $\exists S \in \mathcal{S}_{ik}$ such that $R$ supports $S$, that is, $S \in M(\mathcal{R}'_{ij}, R)$, which contradicts the assumption that $\mathcal{R}'_{ij}$ does not support $\mathcal{S}_{ik}$. A similar argument shows that the reduced set $\mathcal{R}'_{jk}$ does not result in further removals from $\mathcal{R}'_{ik}$. Similarly there can be no unsupported edge $R$ in $\mathcal{R}'_{ij}$ since if there were, one of the removed edges must have supported $R$, since the original clique was stable, and by lemma 4, that edge must have been supported by $R$ and would not have been removed. Hence the order of removals does not matter, every remaining edge is supported, and so the resulting clique is stable. □

## Natural Language Processing

We turn now to briefly describe the NLP pipeline we use for our prototype, which currently forms a model of the user's family relations through dialog, and uses pGMs to track uncertainty and to generate questions. We use three systems: SPLAT, a publically available NLP toolkit [7]; NLPLib, an internal NLP toolkit that uses the averaged perceptron algorithm [8] trained on the part of speech and constituency tree tags and data in OntoNotes Release 4.0 [9], which we used for entity detection [10]; and the Stanford natural language system for labeled dependency trees and coreference resolution [11]. Text is broken into sentences and tokenized. Part of speech tags are added, and the tokens lemmatized. The (labeled) dependency tree is computed. A parse into "chunks" is performed using SPLAT, and named entities are found using NLPLib. Finally the Stanford system is used to generate coreference chains based on the full passage. We then identify nouns, names, pronouns, mentions (which we define as names, pronouns or noun phrases that could be referred to elsewhere in the text), family relations (triggered by tokens such as father, daughter etc.), entities (whose name and gender may or may not be defined at this point), and relation triples, again using simple patterns such as "*My father is named Sam*". The names are identified through combining the named entities identified by NLPLib, 1990 census data, and text patterns, such as *[animate possessive] [is] [named/called] NAME*. Gender identification is done using census data: if a name appears exclusively on the male or female list, it is identified as such; if it is ten times more frequent on one list, it is marked "probably" male/female (this information is used solely to tune the dialog generation); and all other names are left unidentified.

Relation mentions are used to find the relations, taking into account common synonyms. Hypothesized entities are formed from the mentions, and we use the Stanford coreference resolver to group mentions. A special entity is reserved for the narrator of the story, if there is one; gender, name and narrator state are assigned from the mentions, if available. Finally relation triples are formed from pairs of entities together with adjoining relations.

## Z3

The problem of modeling family relations is well suited for first-order logic solvers. We used Z3 to compare this process with using pGMs. Z3 is an open source, SMT-solving theorem prover created at Microsoft[7]. We wrote a set of assertions that governs family relations in general, for example:

*(and (CousinOf a b)  (CousinOf b c))  => (or (SiblingOf a c) (SelfOf a c) (CousinOf a c) (OutOfGraph a c))*

We then wrote a second set of assertions containing the relations that are asserted in the text. Z3 was then used to check that the union of the two sets of assertions is satisfiable. When a solution exists, Z3 returns a satisfying variable assignment; otherwise it declares that none exists.

As it stands, the "out of the box" solution returned by Z3 is not easily interpretable. Many assignments are returned which combine relations with logical operators. Only the first order positive relations [e.g. (*SonOf Jack John*)] are easily interpretable. Furthermore, often the extracted facts lead to multiple possible solutions, of which Z3 returns only one. Additional solutions can be obtained by explicitly excluding solutions found so far, and iteratively ask for another one; but Z3 does not give direct information as to where the uncertainties lie, and so the knowledge returned by the solver could not be used to drive dialog with the user, at least directly. However, Z3 is a powerful solver and we will continue to seek to incorporate it as part of the solution, especially if the logic involved becomes more complex than discussed here.

## Error Correction and Language Learning

Our prototype can correct upstream NLP errors and ambiguities both automatically and with the help of the user. For example, in the sentence *My brother is named Bill and my father is named Bill,* the Stanford systems incorrectly identifies the two *Bills* as referring to the same entity; our system automatically detects that this cannot be the case and corrects the error (note that this is a simple example of a world model being required for coreference resolution; in the sentence *My father is named Bill and my mother's husband is named Bill*, the two *Bills* are coreferent, despite the similar sentence structure). Similar reasoning can be used to resolve ambiguities such as *His* in *John's dad was named Carl. His wife was named Theresa.* In terms of error correction through dialog, an example is as follows: when presented with the sentence *I have a daughter. My daughter's name is Susan.*, the coreference fails to identify *daughter* with *Susan;* the system asks for the name of the daughter, and if told *Susan*, will ask if the two *Susans* are the same person, and if given the answer *Yes* (or equivalent, see below), it will correct the coreference error.

Similar ideas can be used to learn common paraphrases. For example, if the system asks a yes/no question and receives the answer *Indeed!*, then it tells the user that it does not understand *Indeed!,* and asks the same question again. If the user then answers *Yes*, the system has learnt both the answer to the question, and the fact that in the context of a yes/no question, *Indeed!* means the same thing as *Yes.* We have used this idea to gather paraphrase data for the answers to several different question types. It is important however that the context be tracked: for example, the user may write *Susan is indeed my daughter* when asked *Is Susan your daughter?*, which only is equivalent to *Yes* in the specific context of that question. Thus the system would log that if the question is *Is X your daughter*, and the answer is *X is indeed my daughter*, then the latter means *Yes*, for any *X*. Clearly this idea will enable powerful paraphrase learning when crowdsourced in a personal assistant. Some care will need to be taken to weed out spam, but this can likely be done by relying on the fact that most spammers do not collude.

---

[7] Available at: http://z3.codeplex.com/

However, the system's main goal is to reduce uncertainty through dialog. It accomplished this by finding which edge sets in the pGM have more than one element, and then asking the user which relation holds. Clearly this could be made more natural by asking about edge sets that are close in the graph to the narrator's node, and could be made more efficient by finding that edge set whose reduction to a single edge would maximally reduce the collective cardinality of the edge sets throughout the graph.

## Discussion

The main goals of this work were twofold: first, to explore the issues of correctability and interpretability of learned models, using dialog as the main source of new information; and second, to embark on our exploration of using meaning projections in general, to model the semantics of text. The requirement of the explicit, efficient and actionable modeling of uncertainty led us to introduce *possibilistic graphical models*, which provide a simple way to track and update uncertainty to the finest granularity; and to adhere to these desiderata, we have so far avoided using statistical techniques. One very interesting research direction therefore is how to include statistical techniques as needed, in such a way as to maintain these desiderata. The meaning projections we used were low level entity detection, coreference resolution and name detection, and a high level family relations projection. We chose the latter as a first step because it is so well studied and because the relational structure is logical and clear. Of course we are not claiming that modeling family relations from text is itself of research interest (although it may be of practical interest, to automated personal assistants); but this task provided us with a clean test bed for the ideas, and we showed for example that indeed, the family relations projection could be used to correct NLP errors, either automatically or with the help of the user.

Meaning projections can also be largely data driven. As an example, we have built a noun number projection, using Wiktionary. This is a surprisingly nontrivial task. For example, in the sentence *My family is very large, and tonight they are all coming to dinner*, the noun *family* is treated as singular in the first part of the sentence and plural in the second, yet the sentence is grammatically correct. Committing to *family* being either singular or plural can thus lead to coreference errors. We parsed Wiktionary to determine which nouns are definitely singular, which are definitely plural, and which are indefinite. For example, *water*, although non-countable, is singular and has the definite plural *waters*, but *sheep* and *cannon* are both indefinite; and there are plural nouns with no singular form (for example, *People of Spain*), and singular nouns with no plural (for example, *Computer Graphics*), which are also detected. This is a very simple projection, and there are many interesting ways in which large datasets can help with more complex projections. For a particularly elegant example of using the statistics of large datasets to solve what is otherwise a thorny coreference problem, see [12]. We plan to extend these ideas, using for example ClueWeb [13], to help meaning projections to find the patterns they need, automatically.

The main question facing us now is: how to extend the work to include more meaning projections, in such a way that the process is scalable? In fact family relations is one of the more complex examples; one might, for example, model spatial relations with $\mathbb{R} = \{x < 10, 10 \leq x < 1e2, 1e2 \leq x < 1e3, 1e3 \leq x < 1e4, x \geq 1e4\}$ where $x$ is the distance in feet between two entities and bounds are chosen to approximately model $E_1$ being within the {same room, same building, same block, same town, same universe} as $E_2$. Even simple projections could benefit from these ideas: at first blush, the task that a *Possession* projection solves seems clear, but in fact there are many kinds of possession and the inferences one can make will depend on which type one is considering.  If a man possesses a car, and that car possesses a wheel, then it is fair to claim that the man possesses the wheel.  If a man possesses an aunt and the aunt possesses a son, then a similar chain of reasoning does not hold.  It is our hope that the framework presented in this paper will provide a useful tool to help guide these developments.  Eventually, to achieve scalability, the meaning projections themselves will have to be learned from data; perhaps higher level meaning projections could be built automatically from a set of fundamental projections.  The use of meaning projections to automatically build higher level semantic constructs such as scripts and plans [14], which encapsulate typically observed sequences of events, is also an interesting direction for research.

## Acknowledgements

We wish to thank Aitao Chen for generously helping us with his NLPLib package.  We also wish to thank Nicolaj Bjørner for his generous help with Z3.